# Time Series Simulation by Conditional Generative Adversarial Net

Rao Fu[1], Jie Chen, Shutian Zeng, Yiping Zhuang and Agus Sudjianto

Corporate Model Risk Management at Wells Fargo

## Abstract

Generative Adversarial Net (GAN) has been proven to be a powerful machine learning tool in image data analysis and generation [1]. In this paper, we propose to use Conditional Generative Adversarial Net (CGAN) [2] to learn and simulate time series data. The conditions can be both categorical and continuous variables containing different kinds of auxiliary information. Our simulation studies show that CGAN is able to learn different kinds of normal and heavy tail distributions, as well as dependent structures of different time series and it can further generate conditional predictive distributions consistent with the training data distributions. We also provide an in-depth discussion on the rationale of GAN and the neural network as hierarchical splines to draw a clear connection with the existing statistical method for distribution generation. In practice, CGAN has a wide range of applications in the market risk and counterparty risk analysis: it can be applied to learn the historical data and generate scenarios for the calculation of Value-at-Risk (VaR) and Expected Shortfall (ES) and predict the movement of the market risk factors. We present a real data analysis including a backtesting to demonstrate CGAN is able to outperform the Historic Simulation, a popular method in market risk analysis for the calculation of VaR. CGAN can also be applied in the economic time series modeling and forecasting, and an example of hypothetical shock analysis for economic models and the generation of potential CCAR scenarios by CGAN is given at the end of the paper.

---

[1] Email: rao.fu@wellsfargo.com

# 1. Introduction

The modeling and generation of statistical distributions, time series, and stochastic processes are widely used by the financial institutions for the purposes of risk management, derivative securities pricing, and monetary policy making. For time series data, in order to capture the dependent structures, Autoregressive Model (AR), Generalized Autoregressive Conditional Heteroscedasticity (GARCH) Model, and their variants have been introduced and intensively studied in the literatures [3]. Moreover, as an alternative to the time series models, the stochastic process models specify the mean and variance to follow some latent stochastic process, such as Hull White model [4], Ornstein-Uhlenbeck process [4], etc. Similarly, Copula, Copula-GARCH model, and their variants have been applied to capture the complex dependence structures in the generation of multivariate distributions and time series [3]. However, all of these models are strongly dependent on model assumptions and estimation of the model parameters and, thus, are less effective in the estimation or generation of non-Gaussian, skewed, heavy-tailed distributions with time-varying dependent features [3].

Recently, GAN has been introduced and successfully applied in image data generation and learning. GAN is a generative model via an adversarial training process between generator network and discriminator network, where both the generator and discriminator are neural network models [1]. During the GAN training, the discriminator trains the generator to produce samples which resemble real data from some random noise (usually uniform or Gaussian variables) and the discriminator is simultaneously trained to distinguish between real and generated samples. Both neural networks aim to minimize their cost functions and stop once the Nash equilibrium is achieved where none can continue improving [5]. Other variants of GAN have been introduced to improve the training process, such as Wasserstein GAN (WGAN) [6], WGAN with gradient penalty (WGAN-GP) [7], GAN with a gradient norm penalty (DRAGAN) [8] [9], least square GAN (LSGAN) [10], etc. Alternatively, instead of employing an unsupervised learning, Mirza & Osindero [2] proposed a conditional version of GAN, CGAN, to learn the conditional distributions. CGAN performs conditioning into both discriminator and generator as an additional input layer [2] and generates conditional samples based on the pre-specified conditions.

In this paper, we propose to use CGAN to learn the distributions and their dependent structures of the time series data in a nonparametric fashion and generate real-like conditional time series data. Specifically, CGAN conditions can be both categorical and continuous variables, i.e., an indicator or historical data. Our simulation study shows that: (1) Given categorical conditions, CGAN is able to learn different kinds of normal and heavy tail distributions under different conditions with a good performance comparable to the kernel density estimation method, learn different correlation, autocorrelation and volatility dynamics, and generate real-like conditional samples of time series data. (2) Given continuous conditions, CGAN is able to learn the local changing dynamics of different time series and generate conditional predictive distributions consistent with the original conditional distributions.

In terms of application, GAN and CGAN can be appealing alternatives to calculate Value-at-Risk [2](VaR) and Expected Shortfall [3](ES) for market risk management [11]. Traditionally, Historical Simulation (HS) and Monte Carlo (MC) simulation methods have been popularly used by major financial institutions to calculate VaR [11]. HS revalues a portfolio using actual shift sizes of market risk factors taken from historical data, whereas MC revalues a portfolio by a large number of scenarios of market risk factors, simulated from pre-built models. There are cons and pros in both methods. MC can produce nearly unlimited numbers of scenarios to get a well-described distribution of PnL, but it is subject to large model risk of the stochastic processes chosen. While HS includes all the correlations and volatilities as embedded in the historical data, but the small number of actual historical observations may lead to insufficiently defined distribution tails and VaR output [11]. To overcome these difficulties, GAN, as a non-parametric method, can be applied to learn the correlation and volatility structures of the historical time series data and produce unlimited real-like samples to obtain a well-defined distribution. However, in reality, the correlation and volatility dynamics of historical data vary over time, as demonstrated in the large shifts experienced during the 2008 financial crisis. Thus, major banks usually calculate VaR and Stressed VaR as the risk measurements for normal market periods and stressed periods (the 2008 financial crisis period). Thus, CGAN can be applied to learn historical data and its dependent structures under different market conditions non-parametrically, and produce real-like conditional samples. The conditional samples can be generated unlimitedly by CGAN and used as scenarios to calculate VaR and ES as the MC method.

GAN and CGAN can also be applied to economic forecasting and modeling, which has several advantages over traditional economic models. Traditional economic models, such as Macro Advisers US Model (MAUS)[4], can only produce a single forecast for given a condition and under strong model assumptions. By contrast, GAN can be applied as an assumption–free model to produce a forecast distribution given the same single condition. The generation of the forecast distributions can be a useful source of scenarios for Comprehensive Capital Analysis and Review (CCAR) for financial institutions [12].

In Section 2, we formally introduce the GAN, WGAN, DRAGAN, and CGAN, and the proposed an algorithm for CGAN training. We, then, provide an in-depth discussion on the neural network models as hierarchical splines and the rationale of GAN to learn a distribution in Section 3. The simulation studies of CGAN with different distributions, time series and dependent structures are provided in Section 4. We also present two real data studies in Section 5, which includes a backtesting for market risk analysis to demonstrate that CGAN is able to outperform the HS method for the calculation of VaR and ES. We also provide an example of a hypothetical shock analysis of economic models and the generation of potential CCAR scenarios. Section 6 covers further discussion and potential improvement.

---

[2] VaR is the most important measurement in market risk management, which is the assessment of the potential loss of a portfolio over a given time period and for a given distribution of historical Profit-and-Loss (PnL).VAR can be calculated for a pre-specified percentage probability of loss over a pre-specified period of time. For example, a 10-day 99% VaR is the dollar or percentage loss in 10 days that will be equaled or exceeded only 1% of the time (usually over 252 business days or 1 year). In other words, there is a 1% probability that the loss in portfolio value will be equal to or larger than the VaR measurement [11].

[3] ES is another popular measurement of risk in market risk management, as it satisfies all the properties of coherent risk measurement, including subadditivity. Mathematically, ES is the expected loss given the portfolio return already lies below the VaR. Thus, ES gives an estimate of the magnitude of the loss for unfavorable events, but VaR only provide a lower limit of the loss. ES is going to be implemented in the internal models approach (IMA) under the Fundamental review of the trading book (FRTB) framework in around 3 years by major financial institutes [11][13].

[4] The MAUS model is used to generate multiple US macroeconomic forecasts for various applications by financial institutes. More details can be found at *www.macroadvisers.com.*

## 2. GAN & its Variants

### 2.1 GAN

GAN training is a minmax game on a cost function between generator ($G$) and discriminator ($D$) [1], where both $G$ and $D$ are neutral network models. The input of $G$ is $z$, which is usually sampled from a uniform or Gaussian distribution. Formally, the cost objective function of GAN is:

$$\min_G \max_D \; E_{x \sim p_d}[\log D(x)] + E_{\hat{x} \sim p_g}[\log(1 - D(\hat{x}))], \quad (1)$$

where $p_d$ is the real data distribution and $p_g$ is the model distribution implicitly defined by $\hat{x} = G(z)$. Both $D$ and $G$ are trained simultaneously, where $D$ receives either generated sample $\hat{x}$ or real data $x$, and $D$ is trained to distinguish them by maximizing the cost function. While, $G$ is trained to generate more and more realistic samples by minimizing the cost function. The training stops when $D$ and $G$ achieve the Nash equilibrium, where none of them can be further improved through training. In the original paper of GAN [1], the authors also proposed to update the cost function by maximizing the probability of generated samples being real, instead of minimizing the probability of being fake in practice. However, Google Brain conducted a large-scale study comparing these methods, but they did not find any significant difference in performance [14]. Thus, we applied (1) as the original GAN method.

### 2.2 WGAN & DRAGAN

One of the main failure modes for GAN is the generator to collapse to a parameter setting where it always generates a small range of outputs [6]. The literature suggests that a local equilibrium in this minmax game causes the mode collapse issue [8]. Another major issue is the diminished gradient: the discriminator gets too successful that the generator gradients vanish and the generator learns nothing [6]. Two alternative GANs have been introduced to solve these training issues.

On the one hand, WGAN utilizes a new cost function using Wasserstein distance, which enables the generator to improve in any case as the Wasserstein distance between two distributions is a continuous function and differentiable almost everywhere. The cost function of WGAN simplified by Kantorovich-Rubinstein duality is:

$$\min_G \max_{D \in \mathcal{D}} \; E_{x \sim p_d}[D(x)] - E_{\hat{x} \sim p_g}[D(\hat{x})], \quad (2)$$

where $\mathcal{D}$ is the collection of all 1-Lipschitz functions. The literature suggests that the gradient of the discriminator of WGAN behaves better than GAN's rendering the optimization of the generator easier [6]. The Lipchitz conditions are enforced by weight clipping by a small value in the original WGAN paper. Later, Gulrajani et al. [7] proposed to add a gradient penalty in the cost function of WGAN to achieve the same condition instead of using the weight clipping method. However, according to the large-scale study by Google Brain [14] there is no strong evidence showing that WGAN-GP consistently outperforms WGAN after sufficient hyper-parameter tuning and random restarts [14].

On the other hand, DRAGAN is proposed to solve the same issues through a gradient penalty directly to GAN. The authors observed that the local equilibria within (1) often exhibit sharp gradients of the discriminator around some real data points and demonstrated that these degenerated local equilibria can be avoided with a gradient penalty on the cost function. Formally, the cost function of DRAGAN is:

$$\min_G \max_D \{(1) - \lambda E_{\hat{x} \sim p_{d+N(0,c)}}[(\|\nabla D(\hat{x})\|_2 - 1)^2]\}, \quad (3)$$

Kodali et al. [8] showed that DRAGAN enables faster training with fewer model collapses and achieves generators with better performance across a variety of architectures and objective functions.

## 2.3 CGAN

A conditional version of GAN is introduced by Mirza & Osindero in 2014 [2], which enables GAN to generate specific samples given the conditions. CGAN conditions take any kind of auxiliary information from the samples and gives a head start to the generator to create samples. The same auxiliary condition, usually denoted by **y**, are applied to both generator and discriminator as additional input layers. The original input noise variable **z** is combined with **y** in joint hidden representation [2]. Formally, the cost function[5] for CGAN is:

$$\min_G \max_D E_{x \sim p_d}[\log D(x, y)] + E_{z \sim p_z}[\log(1 - D(G(z, y)))], \quad (4)$$

where $p_z$ is the distribution of the random noise, which is usually uniform or Gaussian as the GAN. Similarly, it is easy to make a conditional WGAN (CWGAN) by passing the conditions to the cost function:

$$\min_G \max_{D \in \mathcal{D}} E_{x \sim p_d}[D(x, y)] - E_{z \sim p_z}[D(G(z, y))], \quad (5)$$

CGAN will be degenerated to GAN given a single categorical condition and training CGAN with multiple ordinal categorical conditions is not equivalent to training multiple GANs on each of the conditions, where the former is much harder than the latter in the term of the learning difficulty. It is because the trained CGAN only contains one set of weights/parameters, but able to adjust and generate different distributions given different conditions. On the other hand, CGAN is able to leverage the relationship among different categorical conditions, and effectively utilize the data information across different categorical conditions. Another difference is CGAN can be trained on continuous conditions, which is beyond the reach of using multiple GANs.

---

[5] The cost function for CGAN of the orginal paper [2] is: $\min_G \max_D E_{x \sim p_d}[\log D(x|y)] + E_{z \sim p_z}[\log(1 - D(G(z|y)))]$, which is essentially the same as (4) in practice.

The CGAN described in equation (4) is our main methodology in this paper, and will be tested in the simulation studies under the different conditions described in Section 4. An algorithm for CGAN training is also developed with the consideration of all the major numerical difficulties in the existing literatures. For example, we observed that the mode collapse remains a major issue in CGAN and we currently applied a weight clipping adjustment to remove the sharp gradients of the discriminator, which is in the same spirt of WGAN and DRAGAN to add some gradient penalty on the discriminator. We also found that the algorithm performs better without batch normalization after adding the gradient penalty, which is similar to the suggestion given by Gulrajani et al [7]. We use Adam as the optimization computer as it is the most popular choice in the GAN literature [14]. The neural networks for both discriminator and generator are constructed by fully connected layers followed by LeakyRelu[6] activation functions and implemented in Python Keras Python Deep Learning library [15]. In summary, the algorithm of the CGAN is:

---

Algorithm of CGAN: $n_{dis}$ is the number of discriminator iterations per generator iteration. $c$ the hyper-parameter for weight clipping.

---

**for** number of training iterations **do**

    **for** $n_{dis}$ steps **do**

- Sample a batch from the real data $x$ and their corresponding conditions $y$.
- Sample a batch of noise $z$ from $p_z$.
- Passing $x$, $y$, $z$ to the cost function ((4) for CGAN, (5) for CWGAN) and update the discriminator by ascending its stochastic gradient using Adam.
- Weight clipping $(-c, c)$ to discriminator.

    **end for**

- Sample a batch of noise $z$ from $p_z$.
- Update the generator by ascending its stochastic gradient using Adam.

**end for**

---

[6] Leaky Rectified Linear Unit function (LeakyRelu) with $\alpha$: $f(x, \alpha) = \begin{cases} x, if\ x > 0 \\ \alpha x, \quad else \end{cases}$. LeakyRelu attempts to fix the "Dying ReLU" problem, where a large gradient flowing through a ReLU neuron could cause the weights to update in such a way that the neuron will never activate on any data point again.

## 3. Rationale of GAN

A single layer neural network is a non-linear transformation from its input to its output. A deep neural network can be viewed as a composition of several intermediate single layer neural networks. It is easy to show that a single-layer forward neural network with a ReLU[7] activation function is a multivariate affine spline (also known as piecewise linear spline) [16] and, thus, a deep forward network can be written as a hierarchical multivariate affine splines. Actually, there is a rigorous bridge between the neural network models and the approximation theory via spline functions, and a large class of deep networks models can be written as a composition of max-affine spline functions [17]. In this paper, we work exclusively with network models with forward connected layers followed by ReLU types of activation functions, but the ideas can be generalized naturally to the other types of layers and activation functions.

Let's first focus on the rationale of the simulation of univariate distributions by GAN. In order to generate a real-like distribution, the generator of GAN, a neural network model, is essentially trained by the discriminator to learn the empirical inverse CDF of the training data by spline functions and operators in a non-parametric fashion. For example, let's train a GAN with the $G$ containing a single fully connected layer with 7 nodes followed by a ReLU activation function, and $D$ is constructed similarly. The input of $G$, $z$, followed a U(0, 1) distribution, and the training data, $x$, followed a N(0, 1) distribution. Theoretically, the number of knots of the spline specified by a single layer network is bounded by the number of the network nodes [16]. In this example, the network, $G$, is trained to result in a piecewise linear spline function with 2 spline knots within [0, 1] (the meaningful domain of this mapping) after 10,000 iterations of training, and the other 5 knots are either overlapped at the ones plotted in Figure 1(left) or located out of [0, 1]. It is worth to point out that, here, the training data $x$ is totally independent with input $z$, and the network model cannot be trained by a simple loss function (such as the mean squared error or entropy) for a common supervised learning problem [15]. GAN, as an unsupervised leaning method, provide a non-trivial alternative to learn the same mapping to match the inverse CDF of N(0, 1) from samples randomly sampled from U(0,1), which is due to the help from the $D$ trained simultaneously with $G$.

Next, in order to obtain a smoother spline approximation, we increased the node number of $G$ from 7 to 100 for this example. Batch size and training iterations are increased as well. The result are provided in Figure 1(right), in which the location and number of the spline knots are refined to result in a smoother spline approximation to match the target function. And similarly, most of the spline knots are overlapped at the ones plotted in Figure 1(right) or located out of the support [0, 1]. There is no overfitting in this particular example, and this is expected, as the CDF function of N(0,1) is a monotonic and smooth target function and it should be fitted by a simple spline approximation.

In the previous examples, we demonstrated that GAN with a $G$ containing more nodes of a single layer network results in a smoother spline approximation. Next, we show that using a $G$ containing deeper layers results in a smoother approximation as well. For example, we trained a GAN with the same parameters as the previous example, except that the $G$ containing two fully connected layers with 100 nodes on each layer followed by ReLU. The result is depicted in Figure 2 (left), where a smoother spline approximation is obtained. Here, the usage of networks with multiple layers with ReLU functions is essentially to construct a non-linear mapping by hierarchical splines. Thus, the smoothness of this approximation is expected.

---

[7] Rectified Linear Unit function (ReLU): $f(x) = \max(x, 0)$

*Figure 1 Left: 1-layer with 7 nodes;  Right:  1-layer with 100 nodes.*

*Blue line: inverse CDF from U(0,1) (x-axis) to N(0, 1) (y-axis), Green line: spline approximation trained by GAN.*

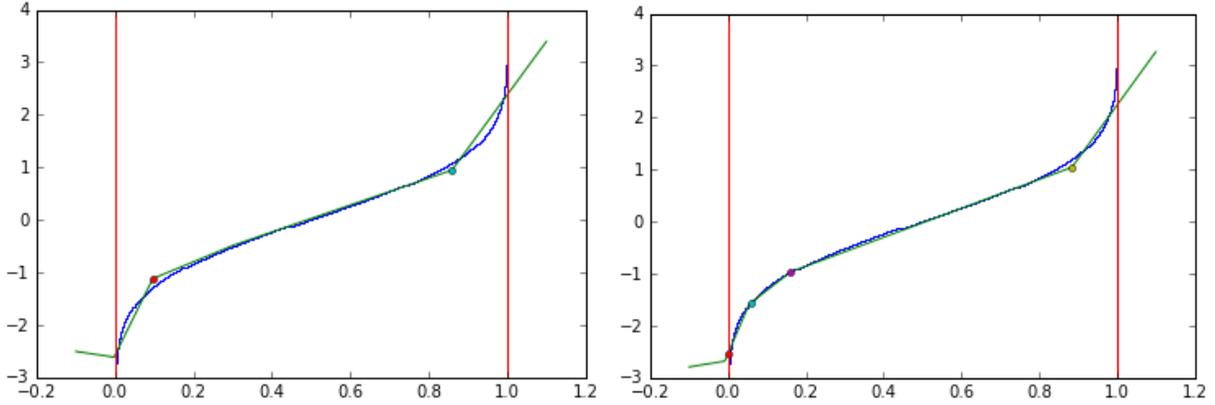

*Figure 2 left: 2-layer with 100*100 nodes with 1 input variable. Right:  1-layer with 100 nodes with 3 input variables*

*Blue line: inverse CDF from U(0,1) (x-axis) to N(0, 1) (y-axis), Green line: spline approximation trained by GAN.*

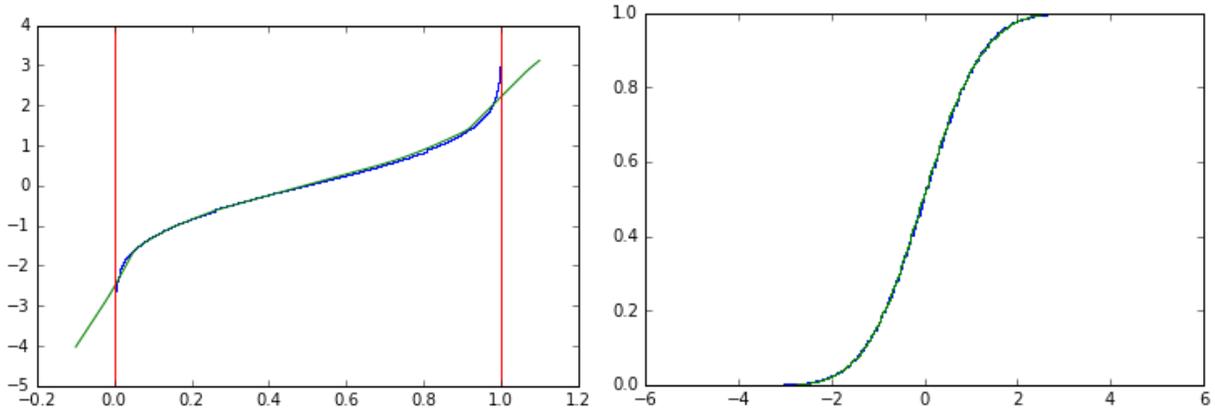

Furthermore, instead of using a single random variable as the input for $G$, $G$ can be sourced from multiple random variables. For example, suppose we decide to learn an univariate N(0, 1) by GAN with 3 independent uniformly distributed random variables as the inputs to $G$, then this is essentially training a $\mathbb{R}^3$ to $\mathbb{R}$ spline to map appropriately from the noise space to the simulated sample space. The inclusion of the multiple inputs to $G$ can also encourage smoothness. For example, we retrained GAN with $G$ containing 1 layer with 100 nodes, but the inputs became 3 independent uniformly distributed random variables. The result is depicted in Figure 2 (right), where we plotted the empirical CDF of the output samples of $G$, which perfectly matched the CDF of the N(0,1).

In terms of the generation of multivariate distributions, the classical method, Copula, usually constructs a complicated mapping from independent uniform variates to the outcomes and depends on strong model assumptions of distributions and parameters to capture the correlation structures. However, GAN provides a straightforward, non-parametric and unsupervised alternative to build the non-linear mapping through neural networks. In section 4, we will present simulation studies to show GAN type method is able to capture the key features of distributions and correlation structures of different multivariate distributions and time series.

In summary, both $G$ and $D$ of GAN, as neural network models, can be written as a composition of spline operators. A smoother spline approximation can be obtained using deep layers, large number of nodes, and multiple random variables as inputs to $G$. CGAN builds non-linear mapping from inputs to outputs with the same rationale as GAN. However, the key difference between CGAN and GAN is that CGAN uses additional variables, which are the conditions given by the user. The condition variables are not independent from the training data. Thus, CGAN is a semi-supervised learning method.

## 4. Simulation Study

In this section, we demonstrate that the performance of the CGAN method on: (1) Mixture of Gaussian distributions with various means and variances, including an example of extrapolation by CGAN. (2) Vector autoregressive (VAR) time series with heavy-tail underling noise and region switching features. The usage of heavy-tail distributions simulates the behavior of real financial data. The region switching time series is a time series with different dependent structures from different time periods/regions, and it is used to simulate the real market movements occasionally exhibiting dramatic breaks in their behavior, which are associated with events such as financial crises. (3) Multivariate generalized autoregressive conditional heteroscedasticity (GARCH) time series with heavy-tail underling noise. Note that both VAR and GARCH models are usually used as parametric models in the financial time series analysis, such as the modeling and prediction of stock rerun, GDP, etc. We show that CGAN, as a nonparametric method, can learn dependence structures of these time series and simulate conditional predictive time series.

To assess the estimation of the heavy tail distributions by CGAN, we use one of the most classic non-parametric methods, Gaussian kernel density estimation, as the benchmarking model, where a cross validation method is used to find the best tuning parameters for the kernel estimation. The estimation is implemented in python scikit learning library [18].

In terms of the architectures of $G$ and $D$ of CGAN, we use a 3-layer forward connected network with 100+ nodes for each layer followed by LeakyRelu ($\alpha = 0.1$) activation function for both $G$ and $D$. The inputs of $G$ are 30 random variables. In this maner, we allow CGAN to have enough complexity and capacity to learn different kinds of multivariate distributions and time series. According to the Google Brain large-scale studies [14], most GAN models can reach similar performance with enough hyper-parameter optimization and random restarts. Thus, we work with CGAN in the following sub-section, where the hyper-parameters are set or trained in the similar setting as the large-scale study from Google Brain [14]. CWGAN is also tested with several examples and results are comparable with CGAN's.

## 4.1 CGAN for Gaussian Mixture Model

We first demonstrate different uses of CGAN to learn and generate a Gaussian mixture distributions. Similar simulation tests have been conducted on T-mixture distributions, and the results are comparable and can be made available upon request.

### 4.1.1 Gaussian Mixture Model with nominal categorical conditions

The CGAN method is first tested on a mixture of Gaussian distributions with various means and variances. Four clusters of 2-dimensional Gaussian distributions (i.e., the training data $x \in \mathbb{R}^2$) with various means and variance are generated (see Figure 3, right panel), where the sample size for each cluster (labelled by different colors) is 1000. The cluster numbers (0, 1, 2, and 3) are the nominal categorical conditions, which are transformed into dummy variables and used in the CGAN training. Thus, the current inputs to $G$ are $z$ and $y$, where $z$ is the random noise and $y$ is the dummy variable categorized for the clusters. In order to learn and simulate the samples for each condition, the CGAN actually trains four non-linear mappings simultaneously, and each non-linear mapping would be applied to generate the samples for each cluster by setting the condition $y$ properly.

In this simulation, CGAN is trained with 10000 iterations and used to generate the four clusters with the same sample size (see Figure 3 left panel). QQ-plots[8] are given to compare the generated distributions and the original ones respectively, and the results show a good match to each other. The benchmarking method, Gaussian kernel estimation, is applied to each cluster, and the generated data by kernel method is compared to the real data. The benchmarking results (Figure 4) show that CGAN and kernel density estimation method have comparable model performance in this example

*Figure 3  2-dimensional Gaussian Mixture distributions: CGAN generated data (left) vs. real data (right). X-axis is for dimension 1 and y-axis is for dimension 2.*

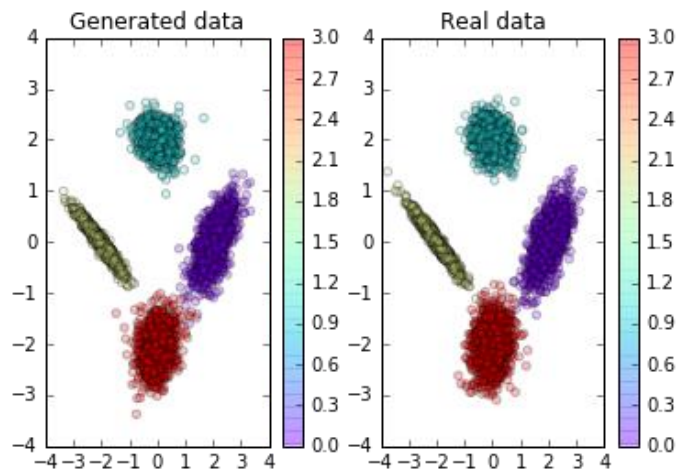

---

[8] The QQ-plots are comparing the 1st dimension of the generated distributions and the original ones (also applied to all the QQ-plots in the following context). The results of the comparison of the 2nd dimension are comparable.

Figure 4 Left: QQ-plot comparing real data vs. CGAN generated data for cluster 1. Right: QQ-plot comparing real data vs. by Kernel estimation generated data for cluster 1. (The results for the other clusters are comparable.)

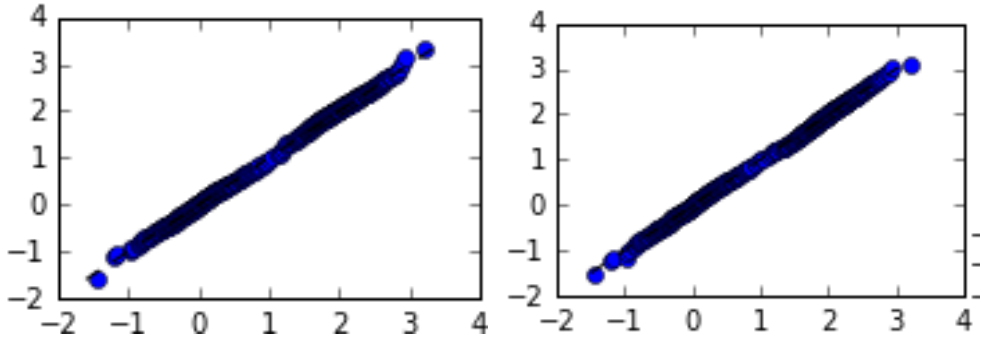

### 4.1.2 Gaussian Mixture Model with ordinal categorical (integer) conditions

In the previous sub-section, the conditions work as the labels of the clusters and are transferred into a dummy variable before passing into the CGAN training process. Alternatively, we can pass the label numbers (0, 1, 2 and 3) directly into CGAN. Thus, the inputs to $G$ are $\mathbf{z}$ and $y$, where $y$ is an integer variable. However, there is no unique labelling of the conditions by integers, and we find that a comparable scales between the data and integer conditions usually give efficient and robust results based on our experiments. Therefore, we recommend rescaling both the training and conditioning data sets properly before the GAN learning process, or always using dummy variables as inputs to nominal categorical conditions.

However, an advantage of using integer conditions is extrapolation. Specifically, we can train a CGAN with integer condition $y$ and feed decimal condition values into the CGAN when generating the conditional samples. For example, we use CGAN to learn the clusters in Figure 3 (right panel), and generate the original clusters and extrapolate the expected clusters between the original clusters (Figure 5, left panel), where the extrapolated clusters located properly between the original clusters with reasonable shapes. Especially, the cluster in Figure 5 (right panel) has a mixed shape between the original clusters. This implies that the trained $G$ of CGAN provides a nonlinear mapping from noise space to simulated sample space that contains a continuous information about the location, shape, mean, variance, and other statistics to generate and extrapolate conditional samples for different clusters.

*Figure 5: Examples of the extrapolation clusters from the original 4 clusters*

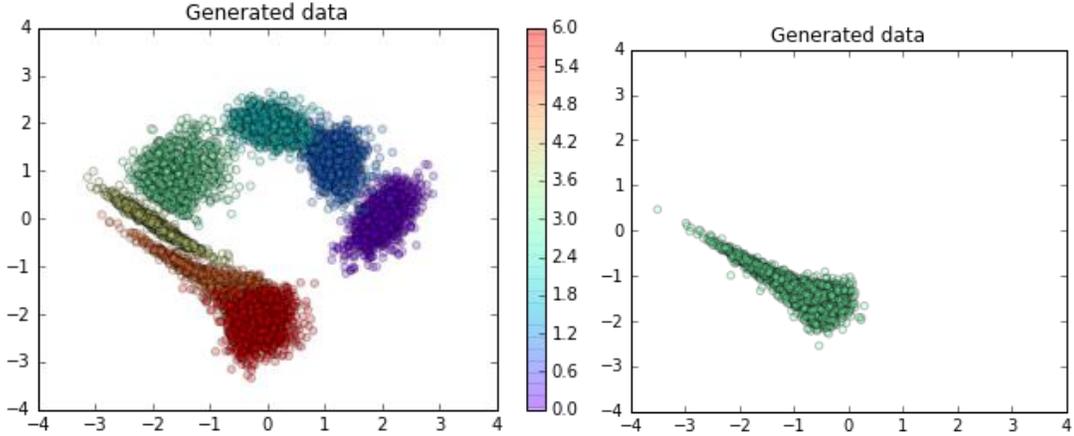

### 4.1.3 Gaussian Mixture Model with Continuous Conditions

Next, we assume each data point of the training data set is sampled from homogeneous distributions with key features varying smoothly. Therefore, in order to capture the varying features, we use CGAN with continuous conditions. Again, as we discussed in 4.1.2, both the data and the conditions would be rescaled if their original scales are not comparable. In the following sub-section, we will first assess the CGAN training and generation on continuous conditions and then assess the extrapolation properties using CGAN by setting the continuous conditions accordingly.

Firstly, the training data for CGAN is generated by Gaussian distributions with varying means and variances for each data point: (1) Generate the means, which are 1000 separated points along the circle with center at (0, 0) and radius = 2. The means are going to be used as the continuous conditions in CGAN training. (2) Generate 1000 variances corresponding to each mean, which are linearly increasing along the circle in an anticlockwise direction. (3) The final data is generated by Gaussian distribution with its mean along the circle and its corresponding variance (Figure 6).

CGAN is trained to learn this synthetic distribution, given the 1000 means as continuous conditions in the training. The outputs are disclosed in Figure 6, and a QQ-plot is drawn to compare the generated distribution with the original one. The results clearly imply that CGAN successfully learns the dynamics in the data and is able to generate real-like samples.

Secondly, similar to the CGAN with integer conditions, one of the most important advantage of CGAN with continuous conditions is extrapolation. In the first example, the dynamics of the changing of variance are indirectly passed into CGAN from the continuous conditions (their means). Thus, we can use CGAN to extrapolate samples beyond the original support, if we feed the CGAN the conditions beyond the original support.

*Figure 6 left: CGAN generated data vs. real data. Right: QQ-plot: comparing real data vs. CGAN generated data.*

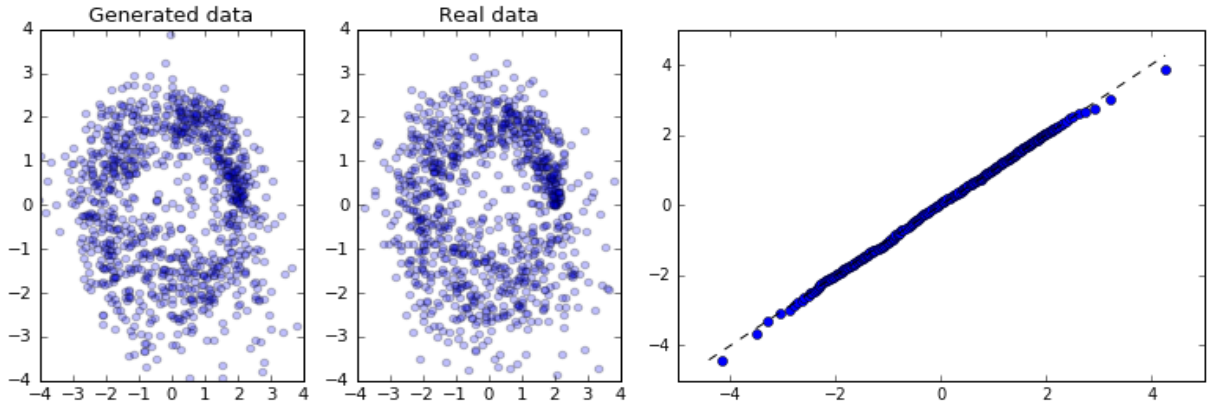

We conduct another simulation study to assess the extrapolation property by CGAN with continuous condition. The training data is generated following the procedure in the previous example except the variance is changing along the x-axis but not a circle. In Figure 7 (right panel), the entire 1000 training data was generated accordingly, where the variance of each data point increases from the original point to both directions along the x-axis. Next, the first and last 20% of the 1000 data were clipped during the CGAN training and CGAN is trained on the remaining 60% data with their corresponding 60% means as conditions (Figure 7, middle panel). After the CGAN training, we simulate and extrapolate 1000 samples with the original 1000 means as continuous conditions. Note that the first and last 20% of the generated data are not within the support of the original training data, but extrapolated by CGAN (Figure 7, left panel). Again, QQ-plots are created to compare the complete distribution generated by CGAN and the original whole distribution (Figure 9 left panel). The results clearly evidence that CGAN demonstrates a good performance of extrapolation at the tails of the distribution, where CGAN learns the dynamic in the changing of the variance is further able to extrapolate the samples beyond the original support following the same dynamics.

Furthermore, we conduct a similar simulation, where the variance increases along the x-axis from the original point with a faster speed. The results (Figure 8 and Figure 9 right panel) show that CGAN can learn different magnitudes of the change dynamics and produce real-like conditional distributions. This is very important in terms of practice as we usually can only access the trend of the change dynamics, but not the magnitude and mechanism. Using CGAN, we can learn the changing dynamics on the support of the training data without any parametric assumption and further extrapolate the data to the unknown area with the same dynamics learned from the training data.

*Figure 7 Slow Variance increasing Scenario—Left: 1000 samples generated by CGAN with the original 1000 means as continuous conditions, middle: clipped data used in the CGAN training. Right: the original 1000 samples.*

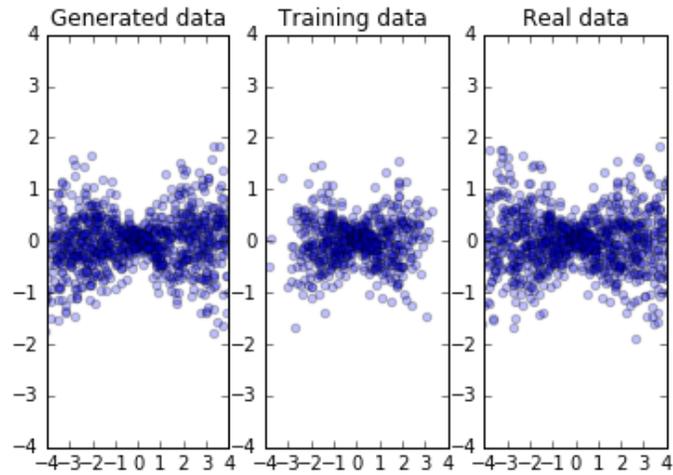

*Figure 8 Quick Variance increasing Scenario: left: 1000 samples generated by CGAN with the original 1000 means as continuous conditions, middle: clipped data used in the CGAN training, right: the original 1000 samples.*

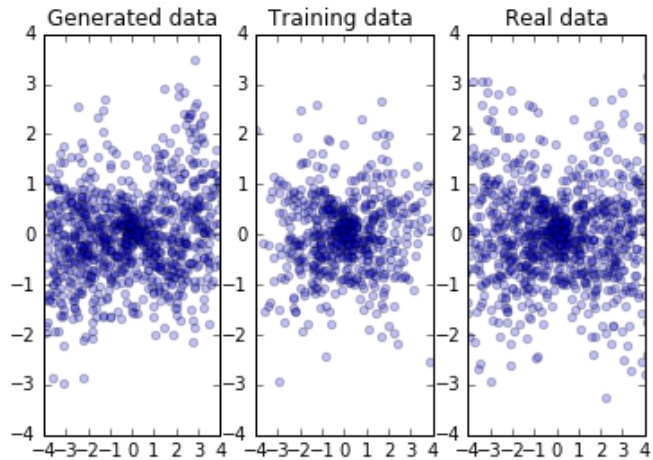

*Figure 9 QQ-plot comparing the generated data vs. real data. Left panel: Slow Variance increasing Scenario, and right panel: fast Variance increasing Scenario:*

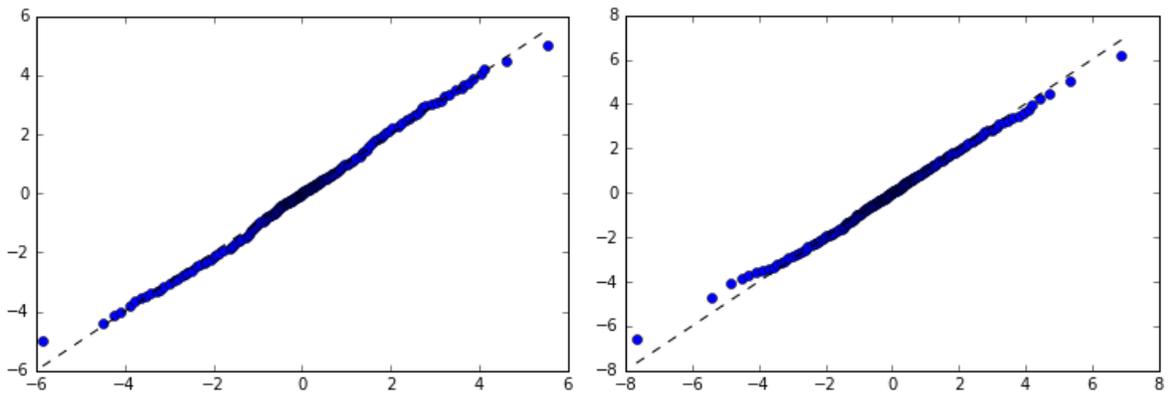

## 4.2 CGAN for VAR Time Series

In the following sub-sections, we extend the usage of CGAN to the learning and generation of time series data. Financial time series analysis is a highly empirical discipline, and numerous classic statistical models have been developed and employed in application, such as AR, ARMA, ARCH, and GARCH [3][1]. In order to have a complete view of the CGAN in time series data modeling, we conduct simulation studies on AR and GARCH-type time series, where the former usually has a strong autocorrelation and the latter has a strong volatility dynamic. VAR model generalizes the univariate AR model, and applied to capture the linear dependent structures among multiple time series. The notation VAR ($p$) indicates a vector autoregressive model of order $p$. Mathematically, the model can be defined as:

$$X_t = c + \sum_{i=1}^{p} a_i X_{t-i} + \varepsilon_t, \qquad (6)$$

where $\{a_i, i = 1, \dots p\}$ are the autocorrelation parameters, $c$ is a constant and $\varepsilon_t$ is an underlying noise. In this sub-section, we generate $\{X_t = (x_{1,t}, x_{2,t}), t = 1, \dots, T\}$, a 2-dimensional VAR (1) time series, as the training data for CGAN modeling. The underlying noises of $X_t$ are sampled from a T distribution, which simulates the heavy tail distributions usually observed in the financial time series. Furthermore, we define the following terminologies as the main statistics to assess the 1st order correlations:

1-lag autocorrelation for time series $i$: $cor_t = \frac{Cov(x_{i,t},\ x_{i,t-1})}{\sqrt{Var(x_{i,t})Var(x_{i,t-1})}}$,

correlation between time series $i$ and $j$: $cor_s = \frac{Cov(x_{i,t}, x_{j,t})}{\sqrt{Var(x_{i,t})Var(x_{j,t})}}$,

cross correlation between time series and time lags: $cor_{st} = \frac{Cov(x_{i,t}, x_{j,t-1})}{\sqrt{Var(x_{i,t})Var(x_{j,t-1})}}$.

Next, we define the 2nd order 1-lag autocorrelation as: $vol_t = \frac{Cov(x_{i,t}^2, x_{i,t-1}^2)}{\sqrt{Var(x_{i,t}^2)Var(x_{i,t-1}^2)}}$. $vol_t$, is used to assess the volatility dynamic of the GARCH time series, where the 2nd order autocorrelation is usually larger than the 1st order. The 2nd order correlation between time series ($vol_s$) and cross correlation ($vol_{st}$) can be defined similarly.

The following sub-section is separated into three parts: 4.2.1 CGAN modeling with continuous conditions, 4.2.2 CGAN modeling with continuous conditions for region switching time series, and 4.2.3 CGAN modeling with categorical conditions for region switching time series.

### 4.2.1 CGAN Modeling with Continuous Conditions

We assume the training data always follows a simple VAR (1) process with the following parameters:

$$X_t = c + aX_{t-1} + \varepsilon_t, \qquad (7)$$

$$c = [0,0]^T, a = [0.8, 0.6]^T, \varepsilon_t \sim T(mean = 0, Cov = \begin{bmatrix} 1 & 0.5 \\ 0.5 & 1 \end{bmatrix}, DF = 6),$$

Note that the underlying noise variable, $\varepsilon_t$, follows a T-distribution with a degree of freedom of 6, which has a much longer tail than a standard normal distribution. The VAR model specifies that the output variables linearly depend on their own previous values and on random noises, thus we propose using the previous 1-time-lag values of the time series as the conditions for the CGAN modeling of the current time series values. The CGAN should be able to capture the dependent structure of VAR (1), and make 1-time-horizon prediction by given the current value as the condition. By this setting, the data format of both the training data and condition data is: Sample size x Number of Time Series (TS) = 1,000 x 2, where the condition data set is constructed by taking 1-time-lag sliding-window snapshot of the training data.

We begin by assessing if the CGAN outputs are truly conditional on their 1-time-lag conditions. Suppose there are two adjacent AR (1) time series outputs $A_{t-1}$ and $A_t$, and the 1-lag autocorrelation between $A_{t-1}$ and $A_t$ is $a$. If the CGAN is able to learn the dependent structure of VAR (1), then the 1-lag autocorrelation between $B_{t-1}$ and $B_t$, generated by CGAN given $A_{t-1}$ and $A_t$ respectively, should be $a^3$ mathematically. Similarly, the 2-lag autocorrelation between $B_{t-1}$ and $B_t$ should be $a^4$. During the CGAN training, we track both the autocorrelations between $B_{t-1}$ and $B_t$ during each iteration, which are the blue (for 1-lag) and green (for 2-lag) lines presented in Figure 10. The horizontal straight lines in Figure 10 represent the corresponding theoretical values of these statistics, which are $a^3$ for 1-lag and $a^4$ for 2-lag. Figure 10 shows that the autocorrelations from the CGAN generated data converge to their theoretical values, however the movement of these statistics is relatively volatile. This is because the sample size of the training data is small, and it can be improved by the increase of sample size (see an example in Figure 16). In summary, the result shows that CGAN is able to learn the autocorrelation and generate a time series with an expected dependent structure. In 4.2.3, we introduce an alternative to generate a time series containing the original dependent structure by using CGAN with categorical conditions.

*Figure 10 Tracking results of autocorrelation (blue curves for 1-lag, green curves for 2-lag) over 10,000 CGAN learning iterations for time series 1 and 2. The horizontal straight lines are the corresponding theoretical values of these statistics.*

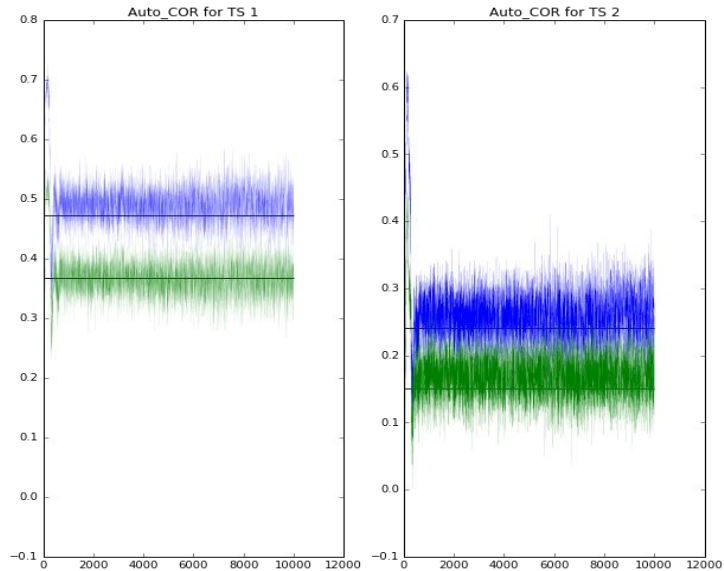

We then assess the 1-time-horizon prediction made by CGAN. The CGAN will be firstly trained given the 1-time-lag values as conditions, and then a conditional distribution given a single condition will be generated by CGAN and compared with the true conditional distribution specified by equation (7). The true conditional distribution is defined through the condition values ($X_{t-1}$), autocorrelation parameters of VAR(1) ($a$) and the underlying noise distribution ($\varepsilon_t$). In the previous section, we already demonstrated that CGAN is able to capture the autocorrelation. Therefore, in order to further test if CGAN can capture the underlying noise distribution and its dynamic, we let the variance of the distribution of $\varepsilon_t$ vary with the sum of the absolute value of 1-time-lag time series values. Formally, the training data is generated by:

$$X_t = aX_{t-1} + \varepsilon_t, \qquad (8)$$

$$a = [0.8, 0.6]^T, \ \varepsilon_t \sim T(mean = 0, Cov = sum(|X_{t-1}|) \cdot \begin{bmatrix} 1 & 0 \\ 0 & 1 \end{bmatrix}, \text{DF}= 20).$$

Here, CGAN was firstly trained with 10,000 iterations, where the training and condition data format is still: Sample size x Number of TS = 1,000 x 2. After the training, 500 conditional 1-time-horizon distributions given 500 random selected conditions were generated and compared with the true ones. The sample size of each generated distribution is the same as 10,000. The distribution comparison is conducted in two way:

(a) QQ-plot: For each condition, we draw a QQ-plot to compare the simulated conditional distribution and the true condition distribution. And we fit a linear regression to assess the fit of the QQ-plot, where the R square and the slope for each comparison are saved and plotted as a histogram plot in Figure 11. There is a clustering at 1 on x-axis for both R squares and slopes, which indicates a good match.

(b) We compare key statistics (mean, variance, skewness, and kurtosis) between the 500 generated conditional distributions by CGAN and the corresponding true conditional distributions. These statistics are plotted in Figure 12 as scatter plots. For example, the means and variances vary in this simulation, and their values from the true conditional distributions (x-axis in Figure 12) are matched along 45 degree line with the ones from the generated conditional distributions by CGAN (y-axis in Figure 12), which implies that CGAN is able to capture these local changes given the conditions. The skewness and kurtosis are fixed in this simulation, which explains why they are located along a vertical line, and the intersection of this vertical line and the x-axis is their true values. The skewness from CGAN is distributed around the true value 0 with a reasonable range of variation compared to the range of the data. The excess kurtosis should be 0 for the normal distribution. But, here, the kurtosis value should be larger than 0 as the underlying noise has a T-distribution with degree freedom of 20. However, there are a few cases where the kurtosis is lower than 0, indicating an underestimation of the heavy-tails, which is unexpected and may not lead to well described tails.

The underestimation of the heavy-tails is due to the insufficient sample size of the training data. Another simulation study was conducted with a much larger sample: Sample size x Number of TS = 20,000 x 2. The new results (Figure 13, Figure 14) show that the 1-time-horizon predication is improved, where the moments from the conditional distributions are all close to the their true values including the kurtosis estimated sufficiently to reflect the heavy-tails. Thus, we conclude that CGAN training is affected by the sample size of the training data, especially for capturing the higher order moments of the distribution.

*Figure 11 histogram of slope (1st column) and R square (2nd column) of the QQ plots comparing distributions generated by CGAN vs. VAR(1). 1st row for the 1st time series (TS), and 2nd row for the 2nd TS.*

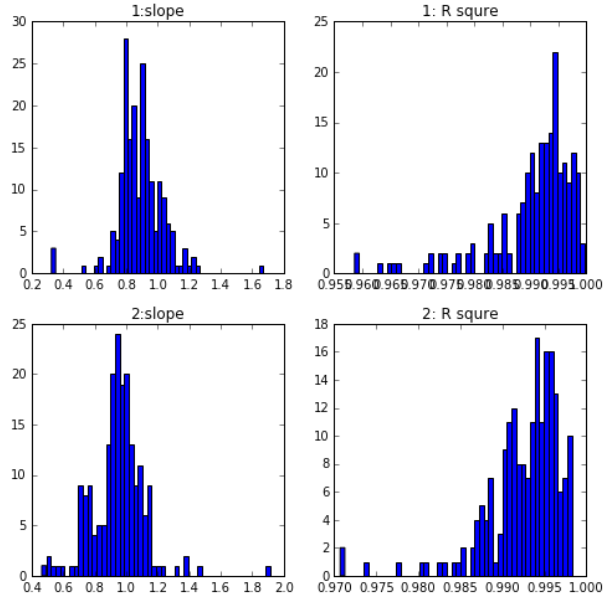

*Figure 12 Mean (1st), variance (2nd), skewness (3rd) and kurtosis (4th) between the CGAN generated samples (y-axis) and real VAR(1) generated samples (x-axis).*

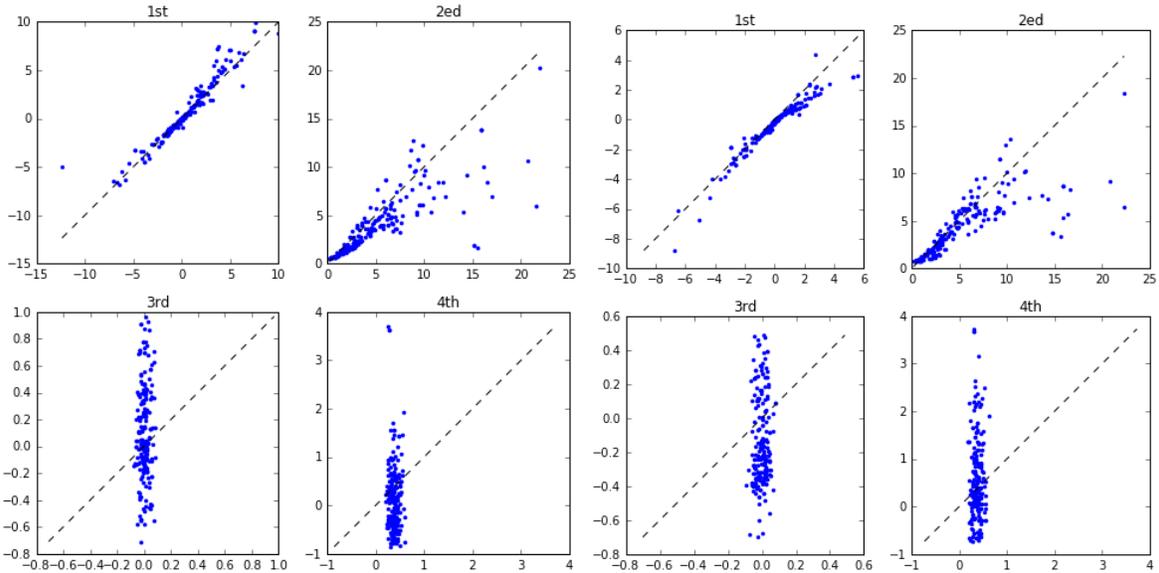

*Figure 13 Larger sample size: histogram of slope (1st column) and R square (2nd column) of the QQ plots comparing distributions generated by CGAN vs. VAR(1). 1st row for the 1st time series (TS), and 2nd row for the 2nd TS.*

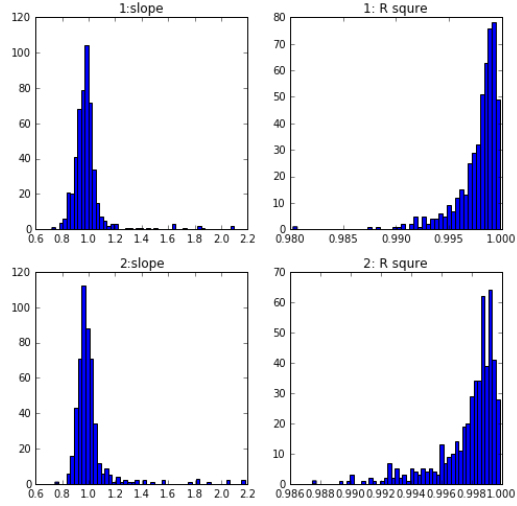

*Figure 14 Larger sample size: Mean (1st), variance (2nd), skewness (3rd) and kurtosis (4th) between the CGAN generated samples (y-axis) and real VAR(1) generated samples (x-axis).*

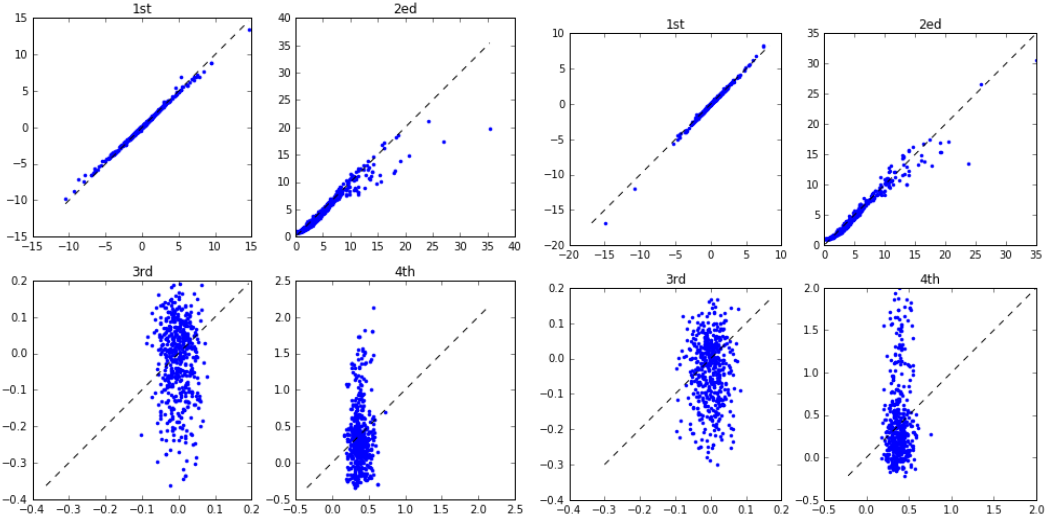

Similarly, we test VAR (1) time series with other volatility changing dynamics, such as:

$$X_t = aX_{t-1} + \varepsilon_t, \quad (9)$$

$$a = [0.8, 0.6]^T, \ \varepsilon_t \sim T\left(mean = \mathbf{0}, Cov = |X_{t-1}|^T \begin{bmatrix} 1 & 0 \\ 0 & 1 \end{bmatrix}, DF = 20\right).$$

The results are comparable, and indicate the CGAN is able to capture different kinds of dynamics of the underlying noise. In summary, we concluded that CGAN is able to learn the dependent structure of the VAR time series and the heavy tails of the underlying noise. CGAN can further generate the conditional 1-time-horizon predictive distributions with the same key statistics as the training data given enough sample sizes.

### 4.2.2 CGAN modeling with continuous conditions for region switching time series

The region switching time series is a time series with different dependent structures from different time periods. As we demonstrated in Section 5.1, both the autocorrelation parameters and the kurtosis of the time series data of equity returns exhibit large changes from the financial crisis period to the normal period. In order to simulate this behavior, we further assume that the time series of the training data follows two different VAR(1) processes in different periods in this simulation. Specifically, the format of the training data is: Sample size x Number of TS = 20,000 x 2, where the first 10,000 x 2 samples are generated given the parameters in region 1, and the rest are generated using the parameters in region 2. Note that region 1 and 2 have different VAR parameters for autocorrelation and underlying noises given below:

Region 1:

$$X_t = c + aX_{t-1} + \varepsilon_t, \qquad (10)$$

$$c = [-1, 0]^T, a = [0.8, 0.8]^T, \varepsilon_t \sim T\left(mean = 0, Cov = \begin{bmatrix} 1 & 0.5 \\ 0.5 & 1 \end{bmatrix}, DF = 6\right),$$

Region 2:

$$c = [1, 0]^T, a = [0.5, 0.5]^T, \varepsilon_t \sim T\left(0, \begin{bmatrix} 1 & 0.3 \\ 0.3 & 1 \end{bmatrix}, 12\right),$$

In this simulation, we follow the same strategy used in section 4.2.1 in which the previous 1-time-lag time series values are the conditions for CGAN training. However, the goal is to test if CGAN can identify the region switch effect: the changes in the VAR parameters for autocorrelation and underlying noises.

CGAN is trained with 10,000 iterations, and the region switching effect is assessed by a comparison of key statistics by scatter plots as section 4.2.1. However, here, 500 conditions are randomly selected from region 1, and product Figure 15 (left panel); and another 500 conditions are sampled to plot Figure 15 (right panel) for region 2. The plots show that the region switching effect is captured by CGAN, because (1) the change of the VAR parameters are captured, as the means from the generated distributions and the true distributions are matched for both regions. (2) The change of variance and the degree of freedom of the underlying T distribution are captured. For example, the excess kurtosis from region 1 is concentrated at 2, while the one from region 2 is concentrated at 0.75. This result is expected since the degree of freedom of the underlying T distribution for region 1 is 6, resulting in a heavier tail of the distribution compared to the one used in region 2 (degree of freedom is 12).

In conclusion, the result implies that CGAN is able to capture the region switching effect in this example. Here, region 1 and 2 are generally not overlapped (see Figure 17, up-right), thus continuous conditions are sufficient. For more general regime switching cases where there are a large amount of overlapped samples from different regions, we need to use both continuous and categorical conditions in CGAN to capture both the global and local changing dynamics of the training data.

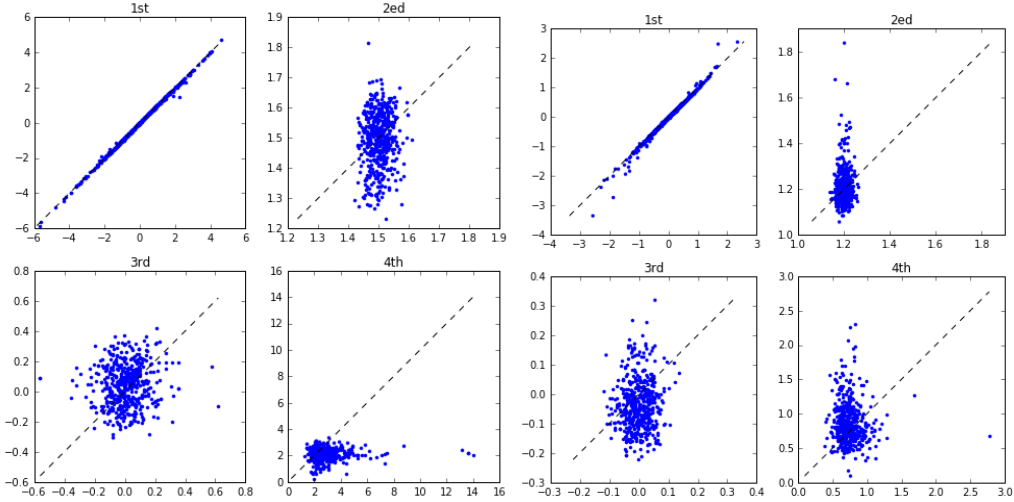

*Figure 15 Assessment of region switching effect: Mean (1st), variance (2nd), skewness (3rd) and kurtosis (4th) from the distributions generated by CGAN (y-axis) vs. VAR(1) (x-axis). Left 2 columns are for Region 1, the rest are for Region 2.*

### 4.2.3 CGAN modeling with categorical conditions for region switching time series

In the previous sections, we used the same strategy as in CGAN training for which a continuous condition is always used to produce a 1-time-horizon prediction. However, in reality, we may not be able to have a continuous condition for each training data, or we are only interested in the generation of unconditional joint distribution of the stationary time series. Also, it is not necessary to assume the distribution of each data point is different. As in the region switching example in section 4.2.2, the data follows the same VAR(1) process within each region. Thus, we can simplify CGAN training by using a categorical variable as the condition for each region, which essentially tells CGAN that samples from different categories have different distributions, dependent structures, and that the same category samples are homogeneous.

However, when CGAN is given a categorical condition, it can only be trained to produce independent and identically distributed (iid) samples. In order to generate time series data with the original dependent structure, we need to reformat the training data by adding a dimension for time. For example, suppose there are only two dimensions of the original training data (Sample size x Number of TS), then we add a time dimension by taking a "T-time-lag" sliding-window snapshot. The updated training data format becomes: Sample size x T x Number of TS. In this way, CGAN is able to learn all of the dependent structure across time and different time series of the training data, and generate real-like time series.

Specifically, in this simulation, the original training data is the same as the one in the previous example specified by (10). As we have previously discussed, this training data is reformatted into "Sample size x T x Number of TS = 20,000 x 2 x 2", where the time dimension is added by taking 1-time-lag sliding-window snapshot. Then, CGAN is trained on the reformatted data set with a categorical variable as the condition for each region. The categorical variable is transformed into dummy variable before using for the CGAN. And CGAN is trained with 10,000 iterations.

We begin by showing that CGAN can generate samples with the same dependence structure as the training data. During the CGAN training process, all the key statistics for dependent structures are tracked. The tracking results are provided in Figure 16, in which the first row shows the tracking results of the correlation statistics, the second row shows the volatility statistics. The first column is for region 1, while the second column is for region 2. The blue curves are the changes of $cor_t$ (in the first row) or $vol_t$ (in the second row) along iteration times, the brown curves are for $cor_s$ (in the first row) or $vol_s$ (in the second row), and the red curves are for $cor_{st}$ (in the first row) or $vol_{st}$ (in the second row). In addition, the horizontal straight lines are the corresponding true values of these statistics in the original training data. In Figure 16, all of the key statistics converge to their corresponding values specified by (10) for different regions, which implies that CGAN output data inherits all the input training data dependent features.

*Figure 16 VAR Tracking results of correlation and volatility over 10000 CGAN learning iterations. X-axis is the iteration time. Y-axis is the correlation/volatility statistics.*

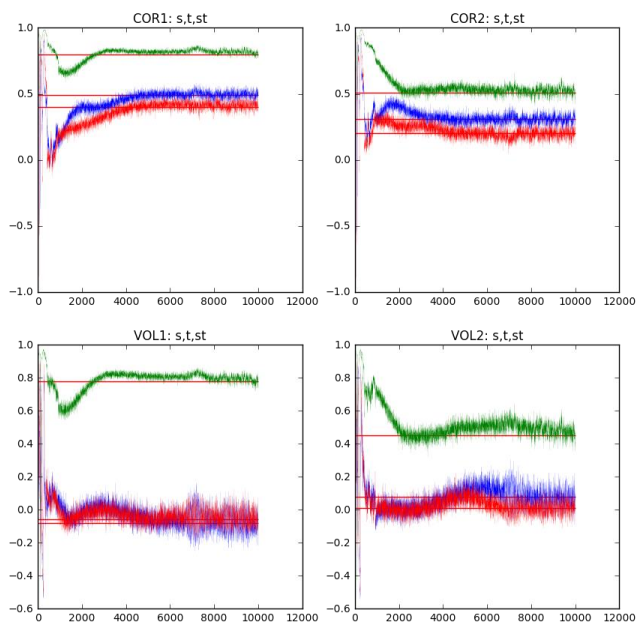

Then, we show that CGAN is able to capture the underlying heavy-tail distribution of the time series. We compare the CGAN generated data with the original data in Figure 17. And QQ-plots are generated to assess the goodness of the fit between the generated and original distributions for both regions, and the results are resonablely well.

In summary, we conclude that CGAN with categorical conditions is able to learn the correlation and volatility dynamics, and the underlying heavy-tail distributions simultaneously. We have also conducted a similar test for VAR(1) with Gaussian underling noise, and we can arrive to the same conclusion.

*Figure 17 Up panel: CGAN generated data (up-left) vs. real data (up-right).*

*Lower panel: QQ-plot for comparing real data vs. CGAN generated data for region 1 (lower-left) and 2 (lower-right).*

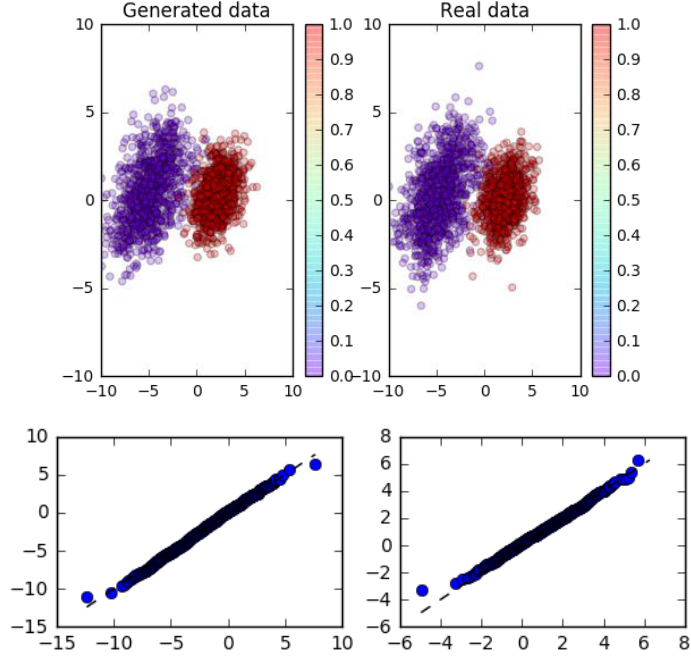

### 4.3 CGAN for GARCH Time Series

In the previous sub-sections, we have demonstrated that CGAN is able to learn the dependent features and the heavy tail distribution of the underlying noise of VAR type time series. In this sub-section, we extend our discussion to the GARCH time series, which usually has a dependent structure in the variance. Formally, a multivariate GARCH (p, q) time series $\{X_t, t = 1, \ldots T\}$ is:

$$\sigma_t^2 = c + \sum_{i=1}^{p} a_i X_{t-i}^2 + \sum_{i=1}^{q} b_i \sigma_{t-i}^2, \quad X_t \sim N(0, \sigma_t^2), \quad (12)$$

where $\{a_i, b_i, i = 1, \ldots p\}$ are the parameters, $c$ is the constant and $\varepsilon_t$ is usually a white noise. As in the VAR model, we use GARCH (1, 1) as an example in the following simulation studies.

We begin with a quick review of the conditions used in the previous sections. For mixture Gaussian/T distributions, we pass a variable containing information of their mean and the changing trend of variance into CGAN. For VAR models, we used the historical 1-time-lag time series values as the conditions, which essentially pass the sample means of the conditional distribution to the CGAN training. However, GARCH time series usually has a small magnitude of the dependent structure of the means, but has a strong correlation in the variance. Thus, the best condition to generate the GARCH time series values at time 't' is the variance at time 't', $\sigma_t^2$. However, the variance information at 't' is usually unknown at time 't-1'. In order to make a 1-time-horizon prediction, we leverage the GARCH volatilities at time 't', $\sigma_t^2$ ($= c + \sum_{i=1}^{p} a_i X_{t-i}^2 + \sum_{i=1}^{q} b_i \sigma_{t-i}^2$) as our "prior" information, and treat them as the conditions for CGAN. However, different from parametric GARCH model, CGAN does not require an assumption of distribution, and can automatically capture the cross correlations among multiple time series. In the future research, we are considering applying the long short-term memory (LSTM) layers [4] into $G$ to extract the historical volatility information automatically. Please note that GARCH volatility $\sigma_t$ only depends on the information up to 't-1', so it is known at 't-1'.

In the following CGAN simulations, besides conditioning on $\sigma_t$, we also show the simulation results conditioning on $\sigma_{t-1}$ as a comparison. Specifically, the training data structure is: Sample size x Number of TS = 10,000 x 2, and the GARCH parameters are:

$$\boldsymbol{\sigma_t^2 = c + aX_{t-1}^2 + b\sigma_{t-1}^2, X_t \sim T\left(0, (\sigma_t^2)^T \cdot \begin{bmatrix} 1 & 0 \\ 0 & 1 \end{bmatrix}, DF = 20\right),} \tag{13}$$

$$\boldsymbol{c = [0.3, 0.3]^T, a = [0.3, 0.3]^T, b = [0.6, 0.6]^T.}$$

CGAN is trained with 10,000 iterations, and we have two CGAN models: CGAN trained conditional on $\sigma_t^2$ (denoted as $\text{CGAN}_{\sigma_t}$), and on $\sigma_{t-1}^2$ (denoted as $\text{CGAN}_{\sigma_{t-1}}$). To assess the model performance, we randomly sample 500 conditions of $\sigma_{t-1}^2$, and use $\text{CGAN}_{\sigma_{t-1}}$ to generate 500 conditional distributions given these conditions. In addition, we calculate 500 $\sigma_t^2$ from the $\sigma_{t-1}^2$ leveraging the GARCH volatilities, and use these $\sigma_t^2$ as conditions to generate 500 conditional distributions by $\text{CGAN}_{\sigma_t}$. Finally, all these conditional distributions either from $\text{CGAN}_{\sigma_{t-1}}$ or $\text{CGAN}_{\sigma_t}$ are compared with the conditional distributions specified by (13). The results are provided in Figure 18, which implies that both $\text{CGAN}_{\sigma_{t-1}}$ and $\text{CGAN}_{\sigma_t}$ are able to learn the changing dynamics of the variance, but $\text{CGAN}_{\sigma_t}$ outperforms the $\text{CGAN}_{\sigma_{t-1}}$ as there is a more clear trend of matching at 45 degree line, demonstrating the importance of using an appropriate condition in CGAN. Also, both of them can capture the heavy tails of the underlying noise, as the kurtosis is concentrated at around a positive value indicating a heavier tail than N(0, 1).

*Figure 18 Mean (1st), variance (2nd), skewness (3rd) and kurtosis (4th) between the generated conditional distributions (y-axis) and the true ones (x-axis). Left panel for $CGAN_{\sigma_t}$ and right for $CGAN_{\sigma_{t-1}}$.*

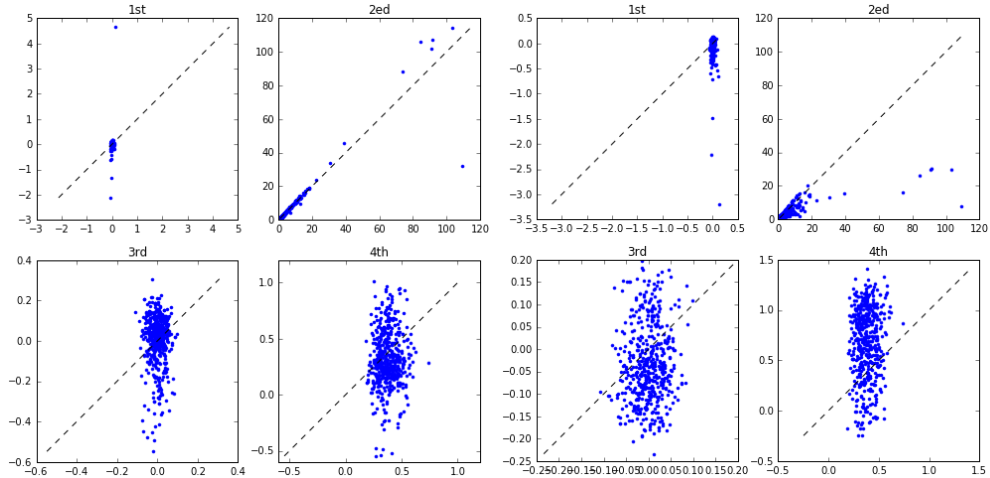

We conduct similar tests on GARCH time series with different parameters and the results are comparable. In summary, we conclude that CGAN is able to learn the dynamics of GARCH time series, the heavy tail of the underlying noise, and generate reasonable 1-time-horizon prediction.

## 5. Real data analysis

In this section, we apply the proposed CGAN method on two real data sets: 1-day stock returns and macroeconomic index data; the former is closer to the GARCH time series and the latter is more like a VAR time series. CGAN is trained by the same algorithm and neural network architectures as the ones in the simulation study. In addition, a backtesting [11] is conducted in 5.1 to show that CGAN is able to outperform the Historical Simulation method for the calculation of VaR and ES.

### 5.1 Equity 1-day Return

We download equity spot prices for WFC[9] and JPM[10] from 11/1/2007 to 11/1/2011 from yahoo finance [19]. And then, 1-day absolute returns are calculated and used as training data. As an example, the time series of the 1-day return of JPM is presented in Figure 19, in which the time series experienced a lot of sharp changes during the 2008 financial crisis.

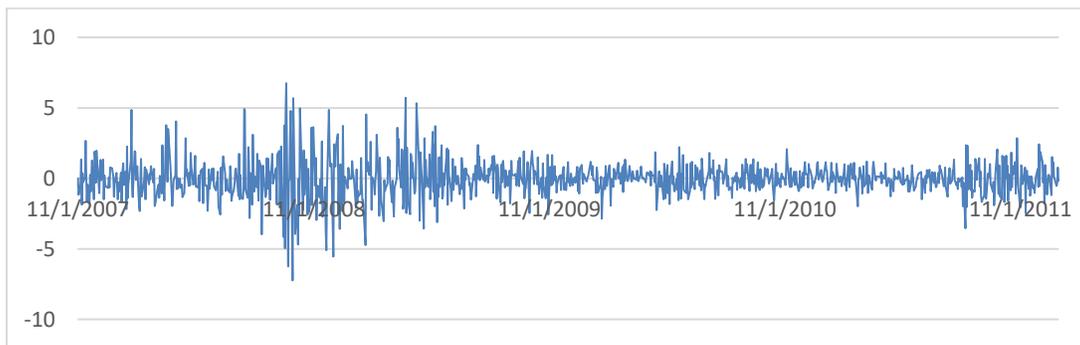

*Figure 19 JPM: 1-day Return Time series*

If we separate the data by crisis period (or the so called stressed period around 11/2007-11/2009) and post-crisis period (the normal period 11/2009-11/2011). We can observe that the joint distribution of WFC and JPM 1-day absolute return has longer tails and larger variances than the ones in the normal period, and the correlation structure is also different (Figure 20, 1$^{st}$ row). This is similar to the region switch time series discussed in section 4.2, where the key data features are changed from time to time.

In market risk analysis, financial institutions usually separate the analysis for the post-crisis period and the crisis period, which leads to the general VaR for normal periods and stressed VaR for stressed periods. By using CGAN, we can easily separate the data learning and generation of the stressed and the normal periods by using an indicator of periods as a categorical condition. In order to generate time series data, we add a dimension to the original data structure as we did in section 4.2.3. The reformatted data structure is Sample size x T x Number of TS = 1,000 x 2 x 2, where the first 500 samples are from the stressed period, and the rest are from the normal period.

The CGAN is trained by 10,000 iterations, and we track the correlation and volatility statistics during the training. Figure 21 shows the tracking results, in which the first row shows the tracking results of the correlation statistics, the second row shows the volatility statistics, and the first column is for condition 1, while the second column is for condition 2. The blue lines are for $cor_t$ (in the first row) or $vol_t$ (in the second row), the brown lines are for $cor_s$ (in the first row) or $vol_s$ (in the second row), and the red lines are for $cor_{st}$ (in the first row) or $vol_{st}$ (in the second row). In addition, the horizontal straight lines are the corresponding true values of these statistics in the training data. All of the testing statistics converge to the true values, but the movement of the second-order correlations is much more volatile than the movement observed in the simulation. This difference takes place since the sample size of the training data is small. In addition, the magnitude of the statistics of this example is smaller than the one in the simulation, and thus hard to capture. After the CGAN learning, we generate the samples for both the stressed and normal periods (Figure 20, 2nd row) and compare them with the real data in QQ-plot (Figure 21).The results show that CGAN is able to learn the dynamics within the data, and generate real-like conditional samples for both periods.

---

[9] WFC: Wells Fargo & Company
[10] JPM: JPMorgan Chase & Co.

*Figure 20 Joint distribution of WFC (X-axis) and JPM (Y-axis) 1-day abs return.1$^{st}$ column is for stress period, and 2nd column is for normal period. Up panel：the real data, and down: the CGAN generated data.*

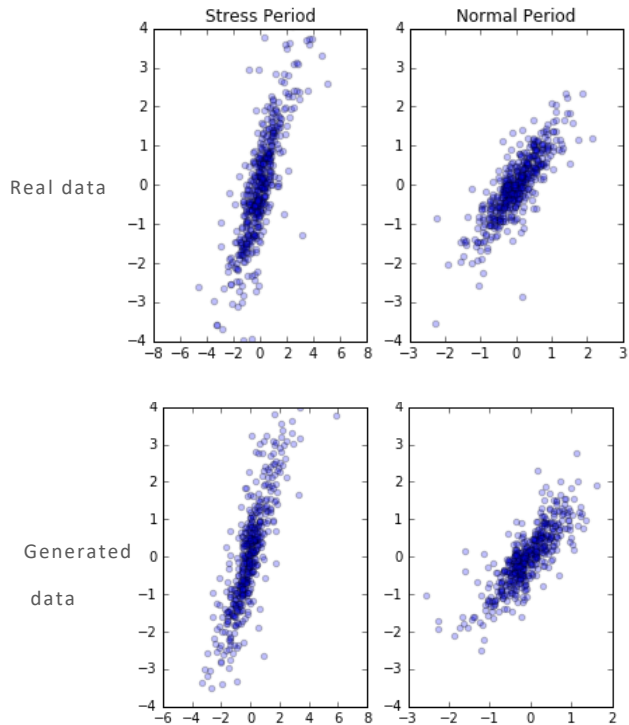

*Figure 21 QQ-plot to compare the real data and the CGAN generated data for stress (left) and normal (right) periods*

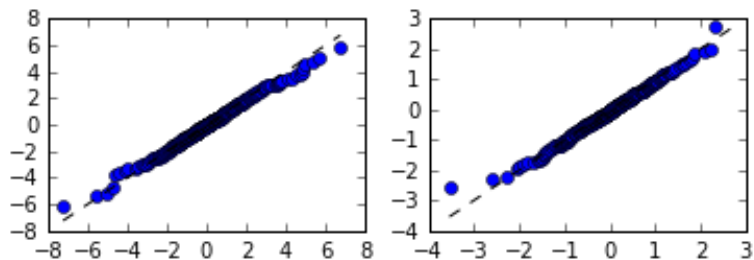

*Figure 22 Tracking results of correlation and volatility parameters over 10000 CGAN learning iterations for equity data.*

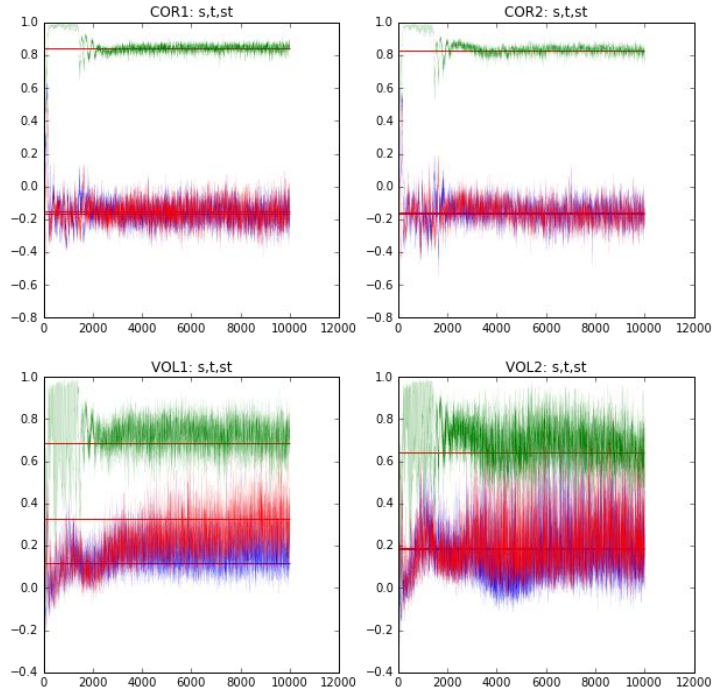

*Table 1 99% 1-day VaR and ES of the testing portfolio calculated by HS and CGAN*

|  | HS method | | CGAN | |
|---|---|---|---|---|
|  | Stress Period | Normal Period | Stress Period | Normal Period |
| VaR (USD) | -4.98 | -2.53 | -7.11 | -3.40 |
| ES (USD) | -6.28 | -3.06 | -9.04 | -4.20 |

As we discussed in Section 1, the Historic Simulation method is one of the most popular methods used by major financial institutions. This method is usually based on a relatively small number of actual historical observations and may lead to insufficient defined tails in the distribution and poor VaR and ES output. However, CGAN can learn the distribution of the historical data and the changing dynamics non-parametrically, and simulate unlimited real-like conditional samples. These samples can be used to fill in the gaps between the original data, which essentially increases the accuracy of the calculation of VaR and ES for market risk analysis. We conduct a backtesting in this example to compare the HS method and CGAN method for the calculation of VaR and ES.

Suppose we have 1 unit of WFC and JPM stock in our testing portfolio. Firstly, the 1-day 99% VaR and ES of this portfolio is calculated by HS for the stressed period (11/2007-11/2009), and the normal period (11/2009-11/2011) which are provided in Table 1. We assume that the data are iid, and the ES is calculated as the average of the tail data values instead of using a weighted average. Next, we plot the left 5% tail of the PnL distributions of this portfolio for both the stressed and normal periods as shown in Figure 23, where the tails have not been sufficiently described by the relatively small number of actual historical data.

Secondly, we use CGAN to learn the historical data for both the stressed and normal periods, and generate a real-like conditional sample set with the sample size 50 times larger than the original one. Then, we calculate the VaR and ES on this larger generated data (Table 1). The plots in Figure 24 show that the large data set generated by CGAN generates a clear tail of the distribution.

Next, additional historical data for WFC and JPM stock prices from 11/1/2011- 11/1/2015 (around 1000 business days) is download from Yahoo Finance in order to implement the backtesting. Since there has been no major financial crisis in this period, we use the VaR and ES from the normal period (in Table 1) as our measurement in the backtesting. The expected breaches over 1000 days for 1-day 99% VaR is 10 days given the iid assumption, but there are 22 breaches using the 99% VaR from HS, while CGAN just results in 8 breaches. In addition, the actual 1-day 99% ES over these 1000 days is -4.04, but ES from HS is only -3.06; while ES from CGAN is -4.20. Table 2 shows that the HS method may lead to an underestimated measurement of the portfolio loss, and CGAN outperformed the HS method in the calculation of VaR and ES for this example.

*Figure 23 left 5% tail of the PnL distribution of the testing portfolio for stress and normal periods*

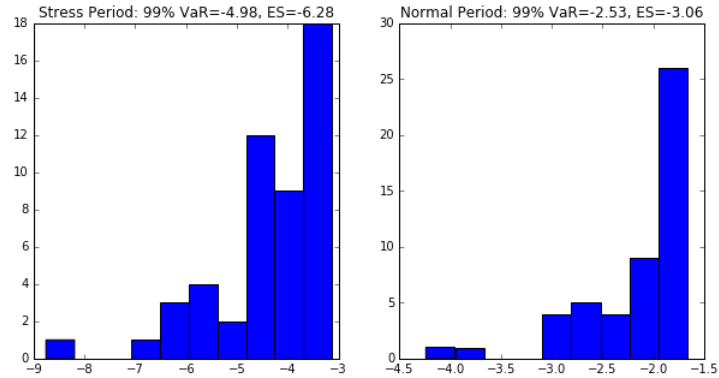

*Figure 24 left 5% tail of the PnL distribution generated by CGAN (with a simple size 50 times larger) for stress and normal periods*

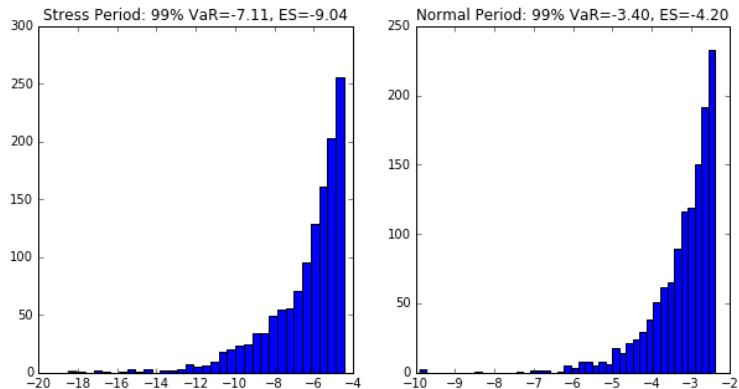

*Table 2 Backtesting results and comparison of 99% ES values*

|  | HS | CGAN |
|---|---|---|
| 99% VaR | -2.53 | -3.40 |
| Breaches over 99% VaR | **22** | **8** |
| Expected breaches at 99% level | 10 | 10 |
| 99% ES | **-3.06** | **-4.20** |
| Actual 99% ES | -4.04 | -4.04 |

## 5.2 Economic Forecasting Model

Furthermore, CGAN can be an appealing alternative in the forecasting and modeling of macroeconomic time series. Large-scale econometric models like MAUS represent the conventional approach, and consist of thousands of equations to model the correlation structures. Each equation is individually estimated or calibrated, and the forecast is produced quarterly and earlier forecasts also serve as inputs for the next quarter forecast. CCAR requires multiple economic forecasts and different capital requirements during different hypothetical economic projections. CGAN-based economic model provides an alternative approach in which multi-quarter forecasts can be produced at once, and the randomness of the output allows the modelers to assess the distributions of the forecast paths.

For example, we downloaded five popular macroeconomic index data from 1956 quarter 1 to 2016 quarter 3 from the U.S. Census Bureau [23]. They are real Gross Domestic Product (GDP), unemployment rate (unemp), Federal fund rate, Consumer Price Index (CPI) and 10-year treasury rate [23]. All these five raw time series are transformed into standardized stationary time series with sample mean 0 and standard deviation of 1. After the transformation, the whole dataset contains 5 variables over 242 quarters. In order to simulate time series data, the initial training dataset (5 variable x 242 quarter) is transformed into the training data for CGAN (230 sample x 5 variable x 13 quarter) by taking a sliding-window snapshot of 13 quarters, just like the operations in the simulation study of section 4.2.3. It is worth noting that the training dataset contains relatively smaller sample size compared to the dimensions of the variable and quarter. Therefore, the training is more challenging and requires a larger number of iterations for each convergence (see Figure 25).

In this section, we use the first four quarters data as the conditions in CGAN modeling to predict the last nine-quarter time series outputs. Thus, the inputs for generator are white noises and the continuous conditions, which is a data set containing 230 sample x 5 variable x 4 quarters. The output is the predictive time series data set containing 230 sample x 5 variable x 9 quarters. The CGAN is trained by 30,000 iterations with a batch size of 100. As in the simulation study, we track the key statistics (mean, standard deviation and autocorrelation) at each iteration to assess the goodness of the predictive samples. In Figure 25, we use GDP as an example, and plot the sample moments and the ones in the training dataset. The first plot shows the convergence of the sample mean around 0. The second plot shows the convergence of sample standard deviation, and the red straight line is the sample standard deviation of the training data. The third plot is the first-order auto-correlation and shows that the model eventually converges to the sample moment.

*Figure 25 Mean, standard deviation and autocorrelation during training for GDP (red line is the value from training data, and blue line is the values from CGAN in each iteration). X-axis is the iteration numbers, and the y-axis is the value of the statistics.*

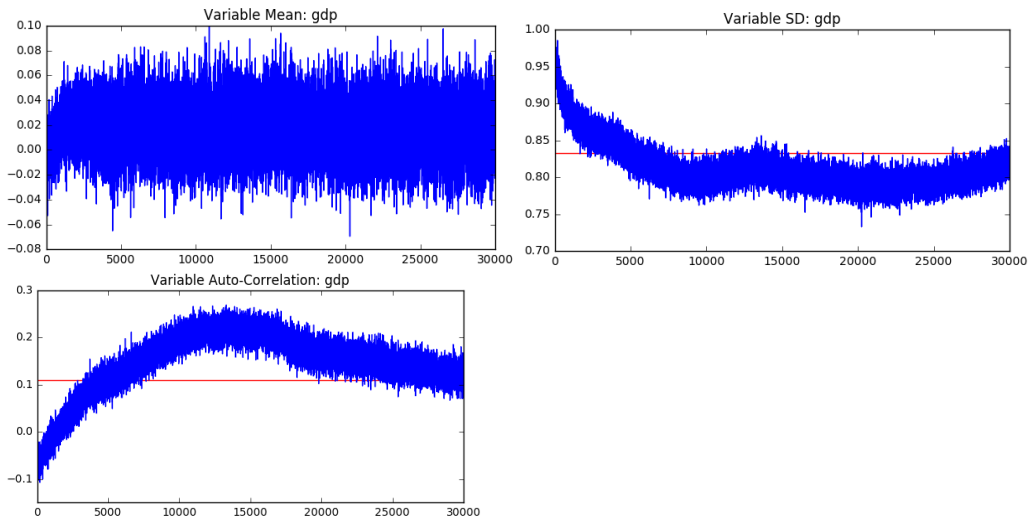

One of the key purposes of economic models is to conduct a hypothetical shock analysis. This kind of analysis is used to construct alternative scenarios for stress testing. In the following example, the Federal fund rate is shocked upward by one standard deviation in the last quarter. The average forecast is generated and compared before and after the shock. A positive shock to the Federal fund rate suppresses the economic activity. Figure 26 (left) shows that the shock leads to a higher unemployment rate during nine-quarter forecast. The Black line is the baseline projection, which is the mean forecast without the shock. The red line is the mean forecast after the shock. The result shows that the shock to the Federal fund rate shifts the unemployment forecast upward on average. Another key advantage of using CGAN for economic forecasting is that the model can produce multiple forecasts and construct an empirical distribution. For example, Figure 26 (right) shows the 100 forecasting paths of GDP generated by CGAN using the most recent four-quarter historical values as conditions. This predicative distribution allows the modeler to assess the likelihood of certain scenarios. For example, the average across all paths serves as the baseline scenario in stress testing, while the bottom 1% path can be used in adverse scenario for stress testing.

*Figure 26 Left: Federal Fund Rate Shock: Unemployment Rate (x axis: quarter, y axis: unemployment rate in percentage). Right: 100 Forecast of GDP Paths Rate (x axis: quarter, y axis: GDP)*

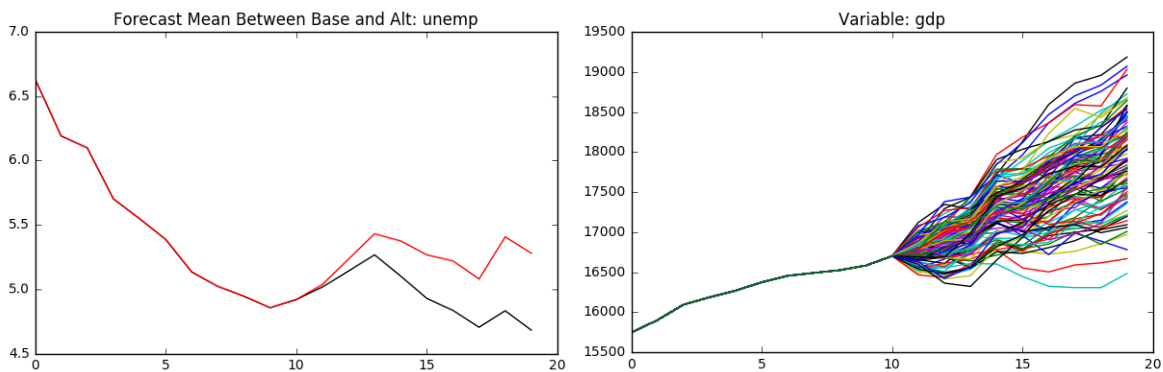

# 6. Conclusion and Future Works

In this article, we propose a CGAN method to learn and generate various distributions and time series. Both the simulation studies and the real data analysis have demonstrated that CGAN is able to learn the normal/heavy tail distributions and difference dependent structures. Moreover, CGAN can generate real-like conditional samples, perform well in interpolation and extrapolation and make predictions, given a proper setting of the categorical or continuous conditions. We consider CGAN an appealing non-parametric benchmarking method to time series models, Copula models, VaR models, and economic forecasting models.

Firstly, regarding future works, more complicated neural network layers can be applied in the discriminator and generator. For example, deep convolutional layers have seen huge adoption in computer vision application. GAN with convolutional networks has been developed, and proved to be an appealing alternative to improve the GAN learning process and perform dimension deduction [20]. In addition, recurrent neural networks and LSTM layers have been recently introduced to formulate volatility and model stochastic processes [4], and LSTM layers have been successfully applied in GAN for stock market prediction on high frequency data [22].

Secondly, instead of using the weight clipping adjustment to the discriminator, we can adopt the gradient penalty method as in the WGAN-GP [7] and DRAGAN [8]. Different cost functions of GAN can be applied as well, such as: LSGAN [10], Non-saturating GAN [14] and Boundary equilibrium GAN [21].

Thirdly, further investigation of the choices of the conditions for CGAN is in the scope of future work. For example, we can train the CGAN conditional on a joint distribution of continuous and categorical variables, which enables CGAN to capture both the global and local changing dynamics of the data. However, by performing this, the learning difficulty gets increases, and we need to involve more advanced neural network structures to build the model.

Finally, more simulation studies can be perform to investigate the ability to capture the second-order correlation of the time series by using CGAN. For instance, we could add the second-order terms in the training data, and enforce the CGAN to learn the first and second order terms at the same time, which can be considered adding more weight of the second-term in the cost function of CGAN. Furthermore, we can use CGAN to learn and generate more stochastic processes, such as QG-models and Volatility surface models; some works have successfully been done by neural network models but not by GAN method [4] [22].

**Acknowledgments**

We would like to thank Vijayan Nair, Richard Liu and Yijing Lu for their consistent support and remarks.